\documentclass[letterpaper, 10 pt, conference]{ieeeconf}  

\IEEEoverridecommandlockouts                              

\usepackage{multirow}

\usepackage[nodisplayskipstretch]{setspace}
\usepackage{graphics} 
\usepackage{epsfig} 
\usepackage{mathptmx} 
\usepackage{times} 
\usepackage{amsmath} 
\usepackage{amssymb}  
\usepackage{algorithm} 
\usepackage{cite}
\usepackage{algpseudocode} 
\usepackage{lipsum}
\usepackage{graphicx}
\usepackage[dvipsnames]{xcolor}
\usepackage{eucal}
\usepackage{mathtools,leftidx}
\usepackage{subcaption}
\usepackage{wrapfig}
\usepackage{bbm}
\usepackage{amsmath}

\usepackage{multirow}

\usepackage{color, colortbl}
\usepackage{xcolor}
\usepackage[colorlinks=false]{hyperref}

\usepackage{bm}

\usepackage[utf8]{inputenc}
\usepackage{fontenc}
\usepackage{siunitx}
\usepackage{tikz}
\usepackage{textcomp}
\newcommand\footnoteref[1]{\protected@xdef\@thefnmark{\ref{#1}}\@footnotemark}

\title{\LARGE \bf
Gaussian Process-based Traversability Analysis for Terrain Mapless Navigation}

\makeatletter
\newcommand*\titleheader[1]{\gdef\@titleheader{#1}}
\AtBeginDocument{%
  \let\st@red@title\@title
  \def\@title{%
    \bgroup\normalfont\large\centering\@titleheader\par\egroup
    \vskip1.5em\st@red@title}
}
\makeatother

\author{Abe Leininger, Mahmoud Ali, Hassan Jardali, and Lantao Liu
\thanks{Abe Leininger is a senior undergraduate student in the Department of Computer Science at Indiana University, Bloomington, IN 47408 USA. {\tt\small aleinin@iu.edu}. Other authors are with  
the Department of Intelligent Systems Engineering at Indiana University, Bloomington. 
{\tt\small \{alimaa, hjardali,lantao\}@iu.edu}.
}
}

\titleheader{ \textcolor{gray}{This paper has been accepted for publication at 2024 IEEE International Conference on Robotics and Automation} \textcolor{blue}{\href{https://2024.ieee-icra.org/}{(ICRA 2024)}} }

\begin{document}

\maketitle
\thispagestyle{empty}
\pagestyle{empty}

\begin{abstract}
Efficient navigation through uneven terrain remains a challenging endeavor for autonomous robots. We propose a new geometric-based uneven terrain mapless navigation framework combining a Sparse Gaussian Process (SGP) local map with a Rapidly-Exploring Random Tree* (RRT*) planner. Our approach begins with the generation of a high-resolution SGP local map, providing an interpolated representation of the robot's immediate environment. This map captures crucial environmental variations, including height, uncertainties, and slope characteristics. Subsequently, we construct a traversability map based on the SGP representation to guide our planning process. The RRT* planner efficiently generates real-time navigation paths, avoiding untraversable terrain in pursuit of the goal. This combination of SGP-based terrain interpretation and RRT* planning enables ground robots to safely navigate environments with varying elevations and steep obstacles. We evaluate the performance of our proposed approach through robust simulation testing, highlighting its effectiveness in achieving safe and efficient navigation compared to existing methods. See the project GitHub\footnote{\url{https://github.com/abeleinin/gp-navigation}} for source code and supplementary materials, including a video demonstrating experimental results.



\end{abstract}

\vspace{-5pt}
\section{Introduction}
In recent years, the utilization of ground autonomous robotics in uneven terrain has significantly expanded, paving the way for a variety of applications in exploration, surveillance, and environmental monitoring. Within this context, the navigation task can be defined as the mission to steer a robot from its current pose to a predetermined goal, optimizing for safety, efficiency, and speed. 
Navigation approaches are typically categorized into either map-based or mapless methods. Map-based strategies require generating an extensive global map of the environment for the robot to plan and navigate through, offering high precision at the expense of substantial computational demands and potential scalability limitations. Conversely, mapless approaches eliminate the need for a global map, instead relying on the robot's immediate sensory inputs for on-the-fly navigation, which sacrifices global optimality in favor of real-time computational efficiency.

In this research, our primary contribution is the construction of a mapless navigation framework composed of two main components. The first one involves utilizing Sparse Gaussian Process (SGP) to create a local map that accurately extracts the geometric characteristics from the robot's immediate surroundings. This process is central to forming a traversability map, which provides the robot with a detailed insight into the navigable parts of the local terrain.
The second component provides a method for footprint assessment and sub-goal selection built on the RRT* local planner for safe and efficient navigation. The RRT* algorithm excels at identifying the most efficient traversable path to local sub-goals or "frontier" nodes, thereby facilitating a gradual progression of the robot toward its global goal.
To validate the efficacy of our approach, we carried out extensive simulation tests on various challenging terrain conditions. These tests confirmed the efficiency of our framework. We also compared our method against a state-of-the-art map-based navigation approach, where we demonstrate a competitive advantage in navigating complex environments.



\begin{figure*} 
\centering
\vspace{5pt}
  \includegraphics[width=0.9\textwidth]{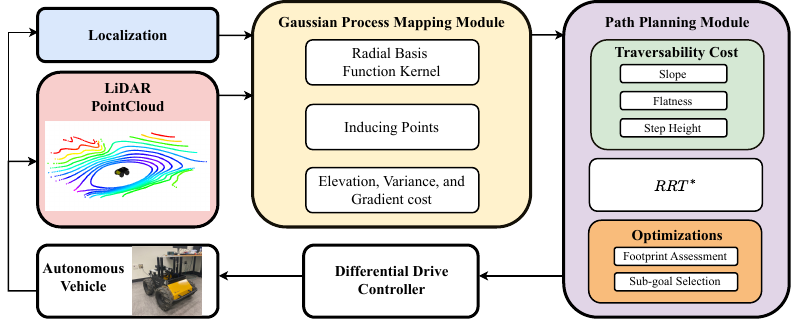}
\caption{\small 
Overview of the proposed uneven terrain navigation framework. From left to right: Localization and LiDAR PointCloud data informs the Gaussian Process Mapping Module. An RBF Kernel trains a model on the induced PointCloud points within this module, yielding Elevation, Variance, and Gradient local cost maps. These maps feed into the Path Planning Module, which constructs a traversability local cost map based on terrain attributes. RRT* planning occurs within the local traversability map and plans a path to the goal. During planning optimizations, including a footprint-based assessment approach and local sub-goal selection. Finally, we send waypoints to the Differential Drive Controller, which sends control commands to the autonomous vehicle.
  \label{fig_overview}
  \vspace{-15pt}
}
\end{figure*}

\section{Related Work} \label{related_work}

Frameworks guiding autonomous systems in navigating uneven terrain generally fall into two categories: geometric-based and learning-based approaches. The former leverages real-time sensor data, primarily from LiDAR and RGBD cameras, while the latter relies on deep or reinforcement learning models\cite{weerakoon2022terp, liang2022adaptiveon}. In this work, we are mostly interested in geometric approaches. These approaches usually construct a map, e.g., an occupancy grid map -- a matrix where each cell represents a spatial area and contains the probability of that area being occupied~\cite{thrun2003learning}. For 3D navigation frameworks, 3D Octomaps \cite{hornung2013octomap},  where the 3D space is subdivided into smaller octants, or the grid-based 2.5D elevation maps, where each grid holds the elevation value \cite{8392399}, are used. Both methods rely on probabilistic approaches. 

Once the map is constructed, a traversability analysis is performed. This analysis can yield a traversability map featuring binary values, indicating whether a grid is traversable or not, or it can employ a cost function to assign a score to each grid \cite{dergachev20222}. 
The traversability score is often calculated by extracting geometric features like slope, roughness, and step height. The slope is determined as the inclination angle originating from a plane fitted to the surrounding terrain, while roughness is measured as the standard deviation of the terrain height values from this plane \cite{5354535}. Additionally, metrics including flatness, sparsity, and unevenness are sometimes considered in the analysis 
\cite{7759199,9738557,5354535,jian2022putn,10160330,s22145217}. 

Following the creation of the traversability map, a global planner finds a path from the robot's current position to the global goal using either graph-based methods such as Dijkstra \cite{dijkstra2022note} and A* algorithms \cite{4082128}, or probabilistic strategies like RRT \cite{rrt} and RRT* \cite{rrts} for trajectory planning \cite{dergachev20222}.  
The standard {RRT}~\cite{rrt} builds a space-filling tree that swiftly explores the state space by randomly sampling states and connecting them to the nearest vertices in the tree. The {RRT*} variant enhances RRT by ensuring optimality, constantly rewiring the tree to find more optimal paths. 
After the path is constructed, the waypoints are sent to a controller.
Alternatively, there are mapless strategies such as DWA \cite{580977}, VFH \cite{borenstein1991vector}, and APF \cite{khatib1986real}, originally designed for 2D navigation, have evolved with advancements in robotics, enabling their use in 3D environments \cite{peng2014obstacle}.
 
Gaussian Process (GP) is a well-established framework to model continuous spatial phenomena\cite{rasmussen2005gaussian,Singh2007,Ouyang2014MAS,ali2023light}. However, despite its proven efficiency, the standard GP comes with a significant drawback which is its high time complexity, $\mathcal{O}(n^3)$. This limitation reduces its applicability in real-time scenarios and with modeling massive datasets.
Addressing this issue, various approaches have been explored to lessen the computational load of GPs. 
One strategy is the Sparse GP (SGP), which reduces the computational complexity of the standard GP by selecting the most relevant subset of the data, $m$ {\em inducing points}, to describe the entire training data using Bayesian rules \cite{snelson2006sparse,sheth2015sparse, titsias2009variational}. 
The computation complexity of SGP, $\mathcal{O}(nm^2)$, is 
typically much smaller than the full GP's computation complexity $\mathcal{O}(n^3)$. In our previous work, we proved that an SGP-based local perception framework is able to navigate the robot in complex cluttered environment \cite{ali2023gp,ali2023exp,mohamed2023gp}.  
Following the paradigm of mapless methods, our approach integrates a Sparse Gaussian Process (SGP) model to build a traversability local map, coupled with an RRT* planner to generate local navigational paths that will lead the robot toward the final goal. 
\section{Methodology} \label{methodology}
We propose a new framework for terrain navigation that combines a GP-based traversability map and an RRT* planner. A systematic overview of the proposed framework is shown in Fig.~\ref{fig_overview}. The framework converts point cloud data into a GP elevation map, which models the terrain geometry around the robot. The prediction of the GP model is then used to analyze the terrain traversability based on three components: local slope, local step height, and neighborhood flatness. The RRT* planner finds a traversable path within the local map while considering the robot's footprint. Finally, we provide an approach for selecting sub-goals while navigating to the global goal. Subsequent sections provide detailed explanations of each module.
\subsection{GP Local Elevation Map}\label{subsec:elevation_map}
In order to construct the traversability map, our initial step involves the creation of a GP-based elevation map, which is built upon the LiDAR observations. Each point in the LiDAR point cloud is represented by a tuple $(x, y, z)$, denoting its spatial coordinates in the 3D space relative to the robot. To ensure that these data points align horizontally to the world, we perform a transformation using the current roll and pitch angles of the robot to align the point's orientation to the global world frame orientation.
This transformation allows us to maintain a consistent geometry analysis with respect to the world frame.
For elevation mapping, each point is decomposed into the point location in the 2D plane, $\mathbf{p}_i=(x_i,y_i)$, and its associated elevation $z_i$.
Consequently, a dataset denoted as $\mathcal D = \left\{\left(\mathbf{p}_{i}, z_{i}\right)\right\}_{i=1}^{n_p}$, is formulated comprising the $n_p$ observed points. 
The proposed elevation local map is a 2D SGP regression model, trained on $\mathcal D$, capable of predicting the elevation value for any point in the continuous space. 
The SGP elevation map can be defined as follows:

\begin{equation}
    \begin{aligned}
    f(\mathbf{p}) &\sim \text{SGP}\left(m(\mathbf{p}), k_{\text{se}}(\mathbf{p}, \mathbf{p}^{\prime})\right), \\
    k_{\text{se}}(\mathbf{p}, \mathbf{p}^{\prime}) &= \sigma_{\text{se}}^2 \exp\left(-\frac{\left\lVert\mathbf{p} - \mathbf{p}^{\prime}\right\rVert^2}{2\ell_{\text{se}}^2}\right),
    \end{aligned}
    \label{eq_mean_kernel_vsgp}    
\end{equation}
where $m(\mathbf{p})$ is the mean function and $k_{\text{se}}\left(\mathbf{p}, \mathbf{p}^{\prime}\right)$ is the Squared Exponential (SE) kernel, also known as the Radial Basis Function (RBF), with a length-scale $\ell_{\text{se}}$ and a signal variance $\sigma_{\text{se}}^{2}$.
In GP regression, a noise $\epsilon_i \sim \mathcal{N}(0,\sigma^2_{n_p})$ is added to the GP prediction to reflect the measurement noise. The elevation $z^*$ for any query point $\mathbf{p}^*$ on the XY plane is estimated by the SGP elevation model as follows:
\begin{equation}
    \begin{gathered} 
    p(z^* | \bm{z}) = \mathcal{N}(z^* | m_{\bm{z}}(\bm{p}^*), k_{\bm{z}}(\bm{p}^*,\bm{p}^*) + \sigma^2_{n_p}), \\
    m_{\bm{z}}(\bm{p})=K_{\bm{p} n_p}\left(\sigma_{n_p}^{2} I+K_{n_p n_p}\right)^{-1} \bm{z}, \\
    k_{\bm{z}}\left(\bm{p}, \bm{p}^{\prime}\right)=k\left(\bm{p}, \bm{p}^{\prime}\right)-K_{\bm{p} n_p}\left(\sigma^{2} I+K_{n_p n_p}\right)^{-1} K_{n_p \bm{p}^{\prime}},
    \end{gathered}
    \label{eq_posterior_mean_kernel_full_gp}
    \vspace{6pt}
\end{equation}
where $\bm{z}=\left\{z_{i}\right\}_{i=1}^{n_p}$, $m_{\bm{z}}(\bm{p})$, and $ k_{\bm{z}}\left(\bm{p}, \bm{p}^{\prime}\right)$ are the posterior mean and covariance functions~\cite{seeger2004gaussian}, 
$K_{n_pn_p}$ is $n_p \times n_p$ co-variance matrix of the inputs, $K_{pn_p}$ is an n-dimensional row vector of kernel function values between $\bm{p}$ and the inputs, with $K_{n_pp} = K_{pn_p}^T$. 
The GP prediction is in the form of a probability distribution with a mean value $\mu_{p_i}$, which indicates the elevation at the query point $\mathbf{p}_i$, and a variance $\sigma^2_{p_i}$, which indicates the uncertainty of this prediction.

To facilitate the comprehensive assessment of the terrain's elevation and slope around the robot, the area covered by the local observation (LiDAR range) is discretized to form a grid, referred to as the local map $\mathcal{M}$. This map is defined by its width $w_m$, height $h_m$, and resolution 
$\gamma_m$. 
Subsequently, the SGP elevation model operates as the analytical tool to assess the terrain's features, i.e. {\em elevation} 
and {\em slope}. 
Each attribute reflects varying facets of the terrain's geometry, thereby providing the build blocks of the local traversability cost map around the robot $\mathcal{M}_\tau$. Additionally, the SGP model provides a confidence level for the predicted elevation represented by the GP uncertainty $\mathcal{M}_\sigma$. We utilize this local map to avoid navigation through uncertain regions.
\subsection{Traversability Map}
\begin{figure*}[th!] 
\vspace{5pt}
\subfloat[Environment\label{fig_trav_a}]{%
  \includegraphics[width=0.2\textwidth,height=0.9in]{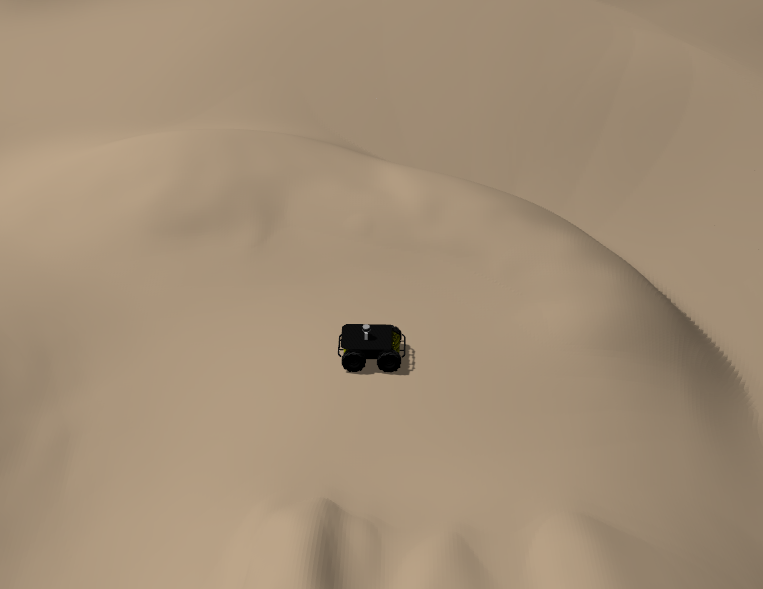}} \hfill
\subfloat[Elevation Map $\mathcal{M}_h$ \label{fig_trav_b}]{%
  \includegraphics[width=0.2\textwidth,height=0.9in]{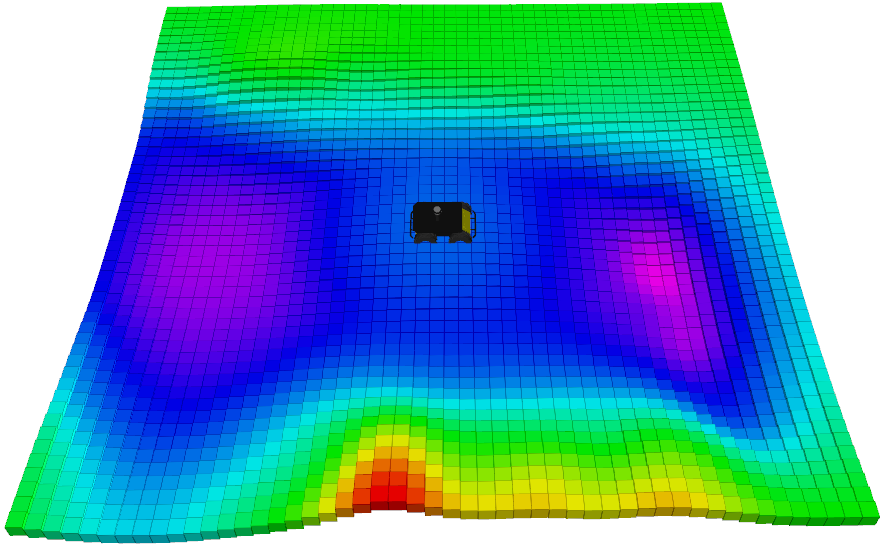}} \hfill
\subfloat[Slope Map $\mathcal{M}_\Delta$ \label{fig_trav_c}]{%
  \includegraphics[width=0.2\textwidth,height=0.9in]{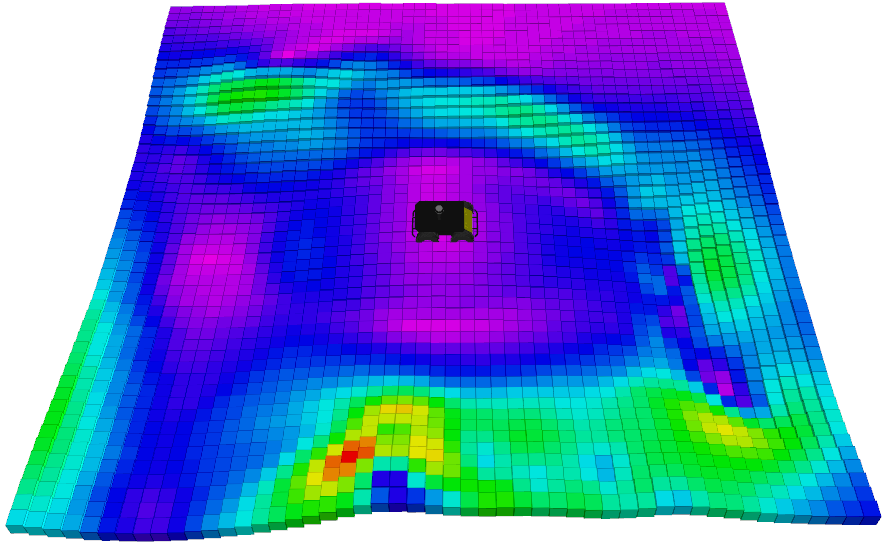}} \hfill
\subfloat[Uncertainty Map $\mathcal{M}_\sigma$ \label{fig_trav_d}]{%
  \includegraphics[width=0.2\textwidth,height=0.9in]{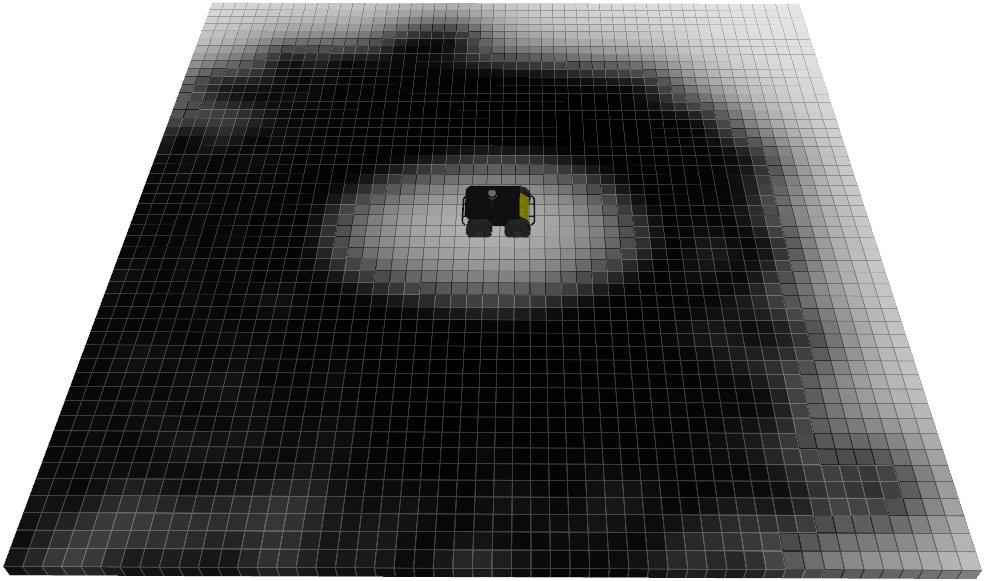} \hfill 
  \includegraphics[width=0.01\textwidth,height=0.9in]{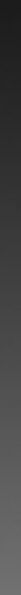}} \hfill
\vspace{4pt}
\subfloat[Flatness $\mathcal{M}_f$ \label{fig_trav_e}]{%
  \includegraphics[width=0.2\textwidth,height=0.9in]{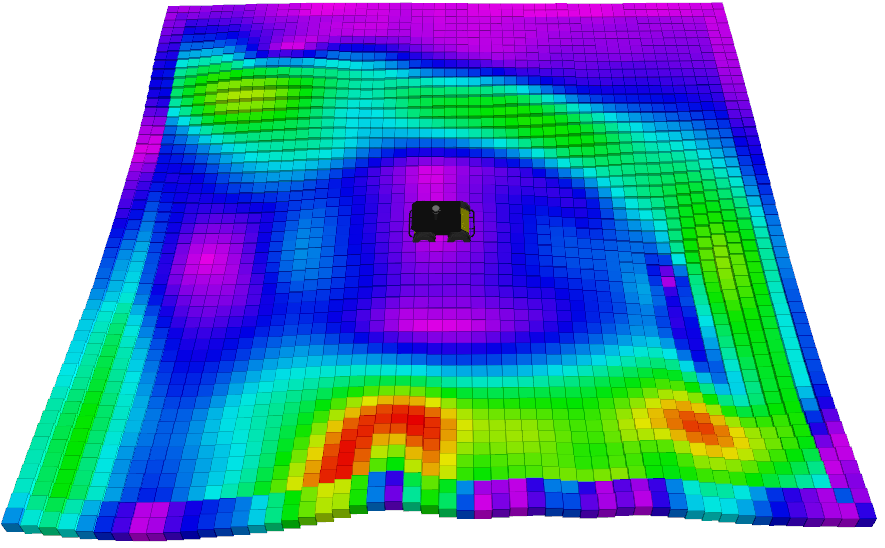}} \hfill
\subfloat[Step Height $\mathcal{M}_\zeta$ \label{fig_trav_f}]{%
  \includegraphics[width=0.2\textwidth,height=0.9in]{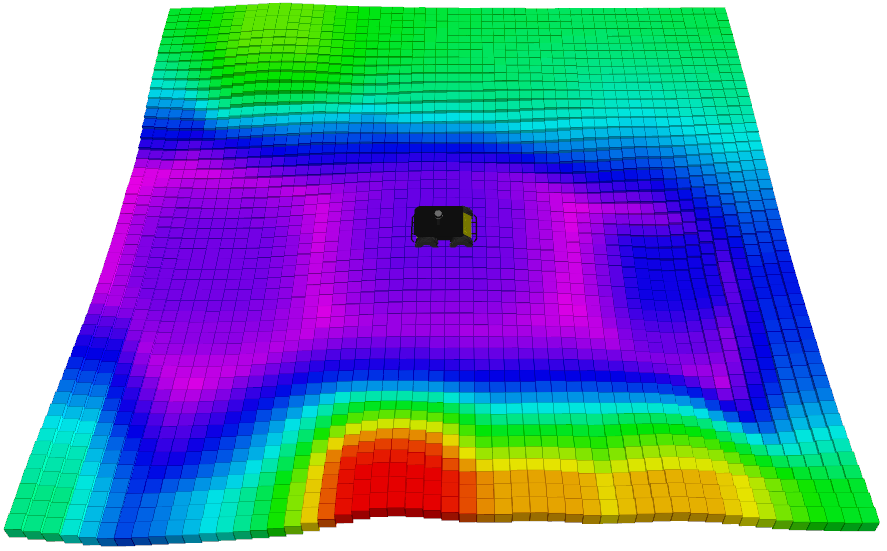}} \hfill
\subfloat[Traversability $\mathcal{M}_\tau$ \label{fig_trav_g}]{%
  \includegraphics[width=0.2\textwidth,height=0.9in]{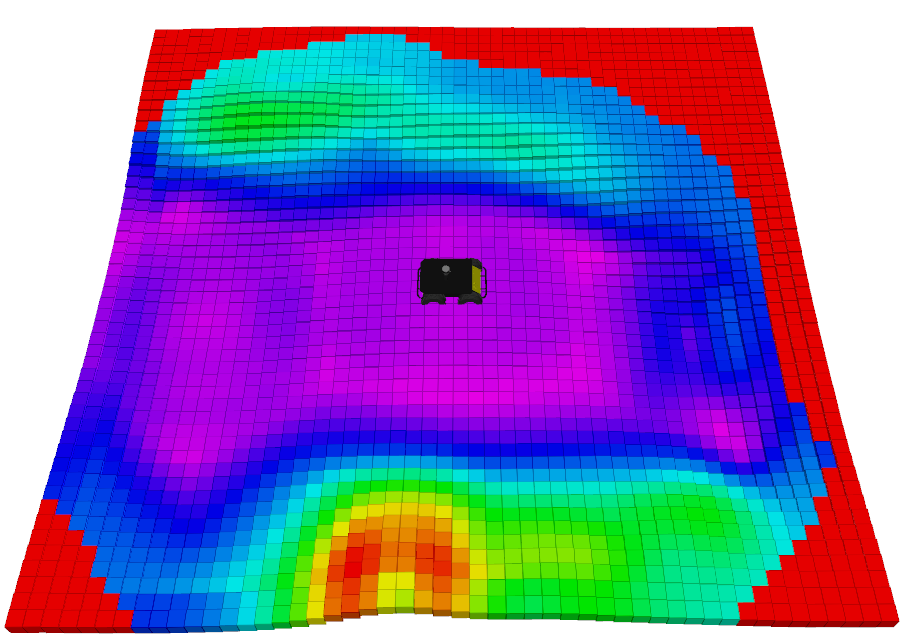}} \hfill
\subfloat[Traversability with RRT\textsuperscript{*}\label{fig_trav_h}]{%
  \includegraphics[width=0.2\textwidth,height=1in]{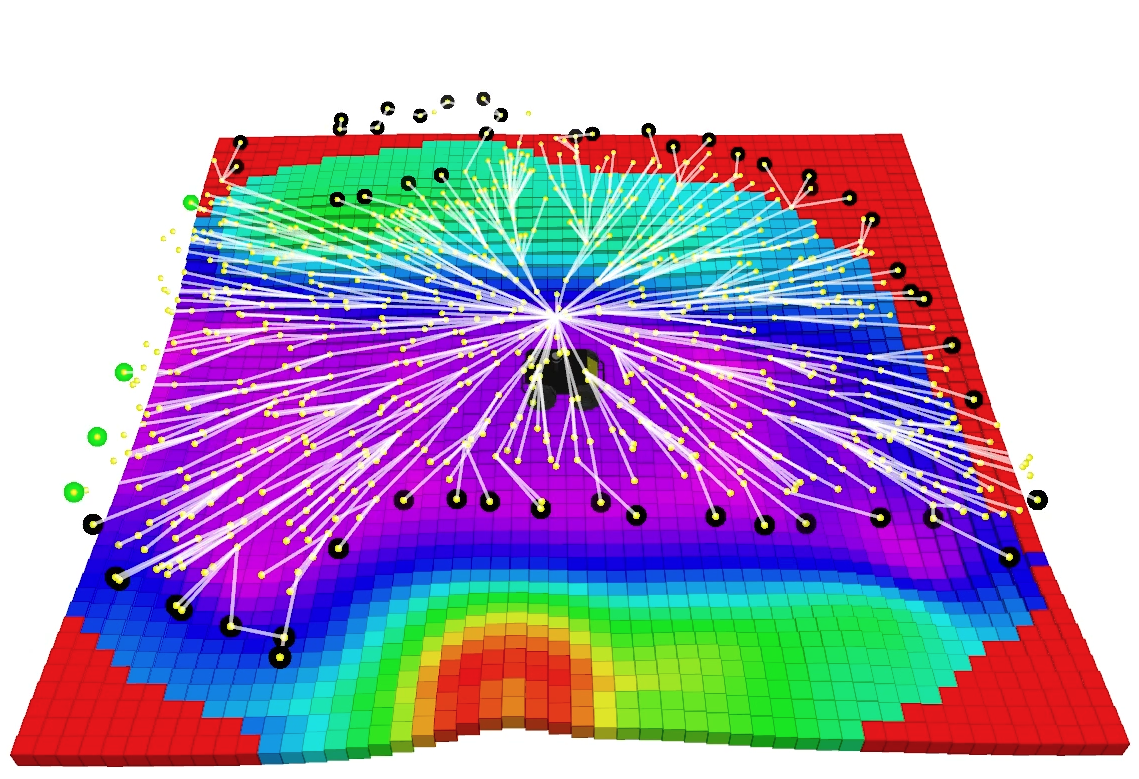}
  \includegraphics[width=0.01\textwidth,height=0.9in]{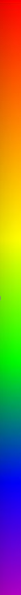}} \hfill
  \vspace*{-2pt}
  \caption{\small Traversability analysis for one observation (marked by a black square in Fig.~\ref{fig_sim_d}) in environment B: (a) shows the simulated environment. (b), (c), and (d) show the local elevation $\mathcal{M}_h$, slope $\mathcal{M}_{\Delta}$, and uncertainty maps $\mathcal{M}_{\sigma}$ respectively, generated by the SGP elevation model. (e) and (f) visualize the flatness $f$ and step height $\zeta$ maps.  (g) shows the final local traversability map $\mathcal{M}_{\tau}$ with the uncertainty map $\mathcal{M}_{\sigma}$ applied as a mask. (h) shows the generated RRT* planning on top of the final traversability map. The traversability values are presented on a rainbow spectrum, with purple representing traversable and red representing non-traversable. Similarly, the Uncertainty map is represented on a grey color gradient where white represents uncertain and black indicating certain regions. 
  \vspace{-15pt}
  }
  \label{fig_trav}
\end{figure*}

To find a safe trajectory to the goal, we build a local traversability map $\mathcal{M}_\tau$. The local traversability cost map characterizes the navigable terrain in the robot's local environment. 
The relative danger is analyzed by assessing terrain characteristics like slope, flatness, and elevation change.
This allows for safe navigation, which adheres to the vehicle's safety limits for roll $\phi$ and pitch $\psi$ angles.
We propose a geometric approach for terrain analysis by utilizing elevation $\mathcal{M}_{h}$ and uncertainty $\mathcal{M}_{\sigma}$ local maps provided by the GP-Module; explained in \ref{subsec:elevation_map}. Also, using the GP-Module, the slope $\mathcal{M}_{\Delta}$ map is generated as the gradient of the SGP model.
Using these provided local maps; we calculate additional terrain feature maps, including flatness $\mathcal{M}_{f}$ and step height $\mathcal{M}_{\zeta}$. Visual depictions of the different local map components are shown in Fig.~\ref{fig_trav}. Together, the slope, flatness, and step height local maps are employed to build a {\em local traversability map} $\mathcal{M}_\tau$, where traversability is set as a value $[0, 1]$, calculated as follows:

\begin{equation}
    \begin{gathered}    
    \mathcal{M}_{\tau} = \omega_1 \frac{\mathcal{M}_{\Delta}}{s_{\text{crit}}} + \omega_2 \frac{\mathcal{M}_{f}}{f_{\text{crit}}} + \omega_3 \frac{\mathcal{M}_{\zeta}}{{\zeta}_{\text{crit}}},
    \label{traversability_evaluation}
    \end{gathered}    
\end{equation}
where \(\omega_1\), \(\omega_2\), and \(\omega_3\) are weights totaling 1, \(s_{\text{crit}}\), \(f_{\text{crit}}\), and \(\zeta_{\text{crit}}\) denote robot-specific critical thresholds for maximum slope, flatness, and step height tolerable before reaching unsafe conditions. These thresholds may be sourced from the robot manufacturer's manual or estimated in advance. 
Additionally, the weighting parameters \(\omega_i\) can be adjusted based on the robot type; for example, a tracked-wheeled robot may perform better on slopes than one with smooth wheels.

Formally, the slope $s$ is calculated as the magnitude 
of the {\em GP-gradient} 
in the local map $\mathcal{M}$:
\begin{equation}
    \begin{gathered}
        s\triangleq \nabla k_{\text{se}}(\mathbf{p}^*, \mathbf{p}_i) = k_{\text{se}}(\mathbf{p}^*, \mathbf{p_i}) \frac{\mathbf{p}_i - \mathbf{p}^*}{\ell_{\text{se}}^2}.
    \end{gathered}   
\end{equation}
The slope matrix $\mathcal{M}_{\Delta}$ is then normalized such that $0 <\mathcal{M}_{\Delta}[i,j]< 1$.
Fig.~\ref{fig_trav_c} visualizes the slope as the color intensity on the elevation map.
Terrain flatness $\mathcal{M}_{f}$ is analyzed by finding the normal vector of the best-fitting plane to a region of points within our local elevation map $\mathcal{M}_h$~\cite{jian2022putn}. 
The region the plane is fit to is equal to the size of the robot footprint $m_\delta$.
We solve the plane equation using the least squares method. 
Step height $\mathcal{M}_{\zeta}$ is calculated similarly to \cite{5354535} by finding the max height within a window relative to the current robot height:
\begin{equation}
    \begin{gathered}
        \mathcal{M}_{\zeta}[i,j] = max(m_{\delta} \in \mathcal{M}_{h}),
    \end{gathered} 
\end{equation}
Finally, the uncertainty map \(\mathcal{M}_{\sigma}\) acts as a {\em mask} over the traversability map \(\mathcal{M}_{\tau}\), filtering out points with uncertainty above a critical threshold \(\sigma_{\text{crit}}\) by marking their traversability as \(\tau = 1\), indicating non-traversable or unsafe areas. This process is formalized as follows: 

\begin{equation}
    \begin{gathered}
        \mathcal{M}_{\tau}[i, j] =
        \begin{cases}
            1 & \text{if } \mathcal{M}_{\sigma}[i, j] > \sigma_{\text{crit}},\\
            \mathcal{M}_{\tau}[i, j] & \text{otherwise.} 
        \end{cases} 
    \end{gathered}
\end{equation}
The mask is visually represented as the red boundary region in the traversability map \(\mathcal{M}_{\tau}\), as illustrated in Fig.~\ref{fig_trav_g}.
\subsection{Path Planning}

Efficient and agile planning modules are essential for generating safe trajectories in uneven environments in real-time. To address this, we utilize the RRT* algorithm~\cite{rrts}, an optimal sampling-based method for motion planning. This method extends the foundational RRT algorithm~\cite{rrt} to include a rewiring step, enhancing its capacity to find more efficient paths by reducing the cost in terms of path length, while also guaranteeing asymptotic optimality.


During RRT* planning a sample point is generated within the  bounds of the traversability map \(\mathcal{M}_\tau\), which is specified as \(p_{\text{new}}(x_n, y_n)\). To position the point in 3D space the elevation map \(\mathcal{M}_h\) provides an elevation value for the point given by \(z_n = \mathcal{M}_h[x_n,y_n]\). We therefore construct a new node \(n_{\text{new}} = (x_n, y_n, z_n) \in \mathbb{R}^3\), effectively projecting \(p_{\text{new}}\) into the 3D space of the local environment.
At this stage, the algorithm must assess whether \(n_{\text{new}}\) qualifies as a traversable leaf node.
Our evaluation leverages a {\em{footprint-based}} assessment approach \cite{7759199}, by considering the robot footprint $m_{\delta} \in \mathcal{M}_{\tau}$. 
When RRT* creates a segment connecting \(n_{\text{new}}\) and \(n_{\ell}\), where \(n_{\ell}\) is the initial branching leaf node.
This segment is then resampled to produce equidistant points, each serving as the center of an analysis area defined by \(m_{\delta}\).
If \(m_{\delta}\), when centered around \(n_{\text{new}}\), lies outside the confines of the local map \(\mathcal{M}\), \(n_{\text{new}}\) will be added to \(V_{\mathcal{F}}\), a vector holding frontier nodes. 
A frontier is defined as an available sub-goal within the current local map.
We analyze the local traversability values inside each $m_{\delta}$ by considering the number of non-traversable cells $n_{\text{coll}}$. We count the number of non-traversable cells within each $m_{\delta}$ that is greater than $\tau_{\text{crit}}$.
\begin{equation}
    \begin{gathered}
n_{\text{coll}} = \sum_{i=1}^{|m_{\delta}|}\sum_{j=1}^{|m_{\delta}|} \mathbf{1} (m_{\delta}[i, j] > \tau_{\text{crit}}).
    \end{gathered}
\end{equation}

\begin{figure} \centering
\includegraphics[width=0.35\textwidth,height=1.6in]{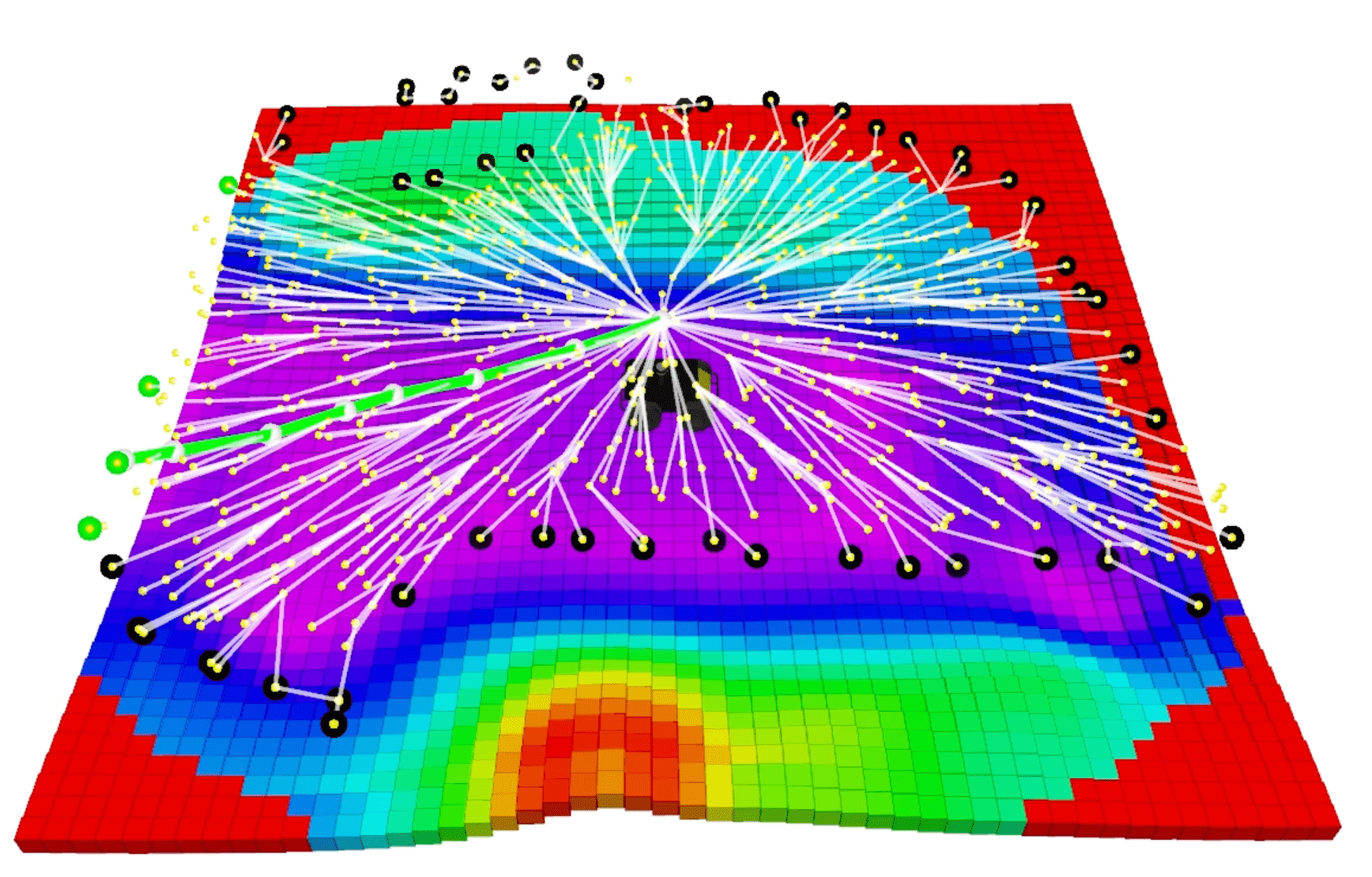} 
  \caption{\small 
  Expanded Fig.~\ref{fig_trav_h} illustrates an RRT* planning iteration, where the green trajectory highlights the route to the chosen sub-goal. Black nodes signify edge nodes leading the untraversable areas. While green nodes indicate frontier nodes and potential targets for subsequent planning iterations.
   \vspace{-10pt}
    \label{trav_planning}
   }
\end{figure}

Notably,~\(\tau_{\text{crit}}\) denotes how aggressive the planning will be within \(\mathcal{M}_{\tau}\). If the collision count, \(n_{\text{coll}}\), is zero in the evaluated \(m_{\delta}\), it is deemed traversable. Consequently, the corresponding node is added to the vertex set of all leaf nodes, \(V_{\ell}\). However, if \(n_{\text{coll}}\) exceeds a predefined threshold  \(\text{coll}_{\text{max}}\), the new node \(n_{\text{new}}\) is identified as an edge node and appended to $V_{\mathcal{E}}$, a vector holding the edge nodes within the local map. 
These nodes will be removed from the \(V_{\ell}\) vector, halting the tree's growth when branching into a non-traversable area. \(\text{coll}_{\text{max}}\), is established based on the number of non-traversable cells in~\(m_{\delta}\). See Fig.~\ref{trav_planning} to view the visual representation of edge and frontier nodes. One benefit of introducing \textit{edge vertices} is that they facilitate avoidance of local minima, preventing the robot from getting stuck.
 
\subsection{Sub-goal Selection}
If the global goal is not attained within the current iteration of the planning algorithm, the branching process persists until a viable sub-goal is identified, to which the robot can navigate and subsequently initiate a re-planning process. It's worth noting the difference between an edge node and a frontier node: an edge node refers to a leaf node leading to a non-traversable area, whereas a frontier node is located at the boundary of the local map, regarded as a potential sub-goal candidate. 
We employ a weighted-sum approach to determine the optimal sub-goal, denoted as \(\mathbf{g}^*\), by evaluating two distance metrics at each frontier node \(n_i \in V_{\mathcal{F}}\). The process is defined mathematically as: 
\begin{equation}
    \begin{gathered}    
    D(n_i) = \text{norm}\left(\sqrt{(G_x - n_{i_x})^2 + (G_y - n_{i_y})^2}\right),\\
    E(n_i) = \text{norm}\left(r - \sqrt{(n_{i_x})^2 + (n_{i_y})^2}\right),\\
    \mathbf{g}^* = \arg \min_{n_i \in V_{\mathcal{F}}}\left(\alpha_1 D(n_i) + \alpha_2 E(n_i)\right).
    \label{weighted_sum_optim}
    \end{gathered}    
\end{equation}

Here, we introduce two normalized distance functions, \(D(n_i)\) and \(E(n_i)\), that are utilized in a weighted sum to find the minimum and thus select the nearest sub-goal. The function \(D(n_i)\) represents the Euclidean distance from the current node $n_i$ to the final goal with coordinates \((G_x, G_y)\) in the robot's frame. Conversely, \(E(n_i)\) measures the distance from the frontier node to the boundary of the local map. For simplicity, we consider the local map as a circle around the robot with a radius $r$ equal to half the map width. The weights \(\alpha_1\) and \(\alpha_2\) serve to balance the influences of the two distance metrics.


\begin{figure} \centering
\vspace{5pt}
\subfloat[\label{fig_sim_a} Gazebo Environment \textbf{A}]{%
  \includegraphics[width=0.24\textwidth,height=1.4in]{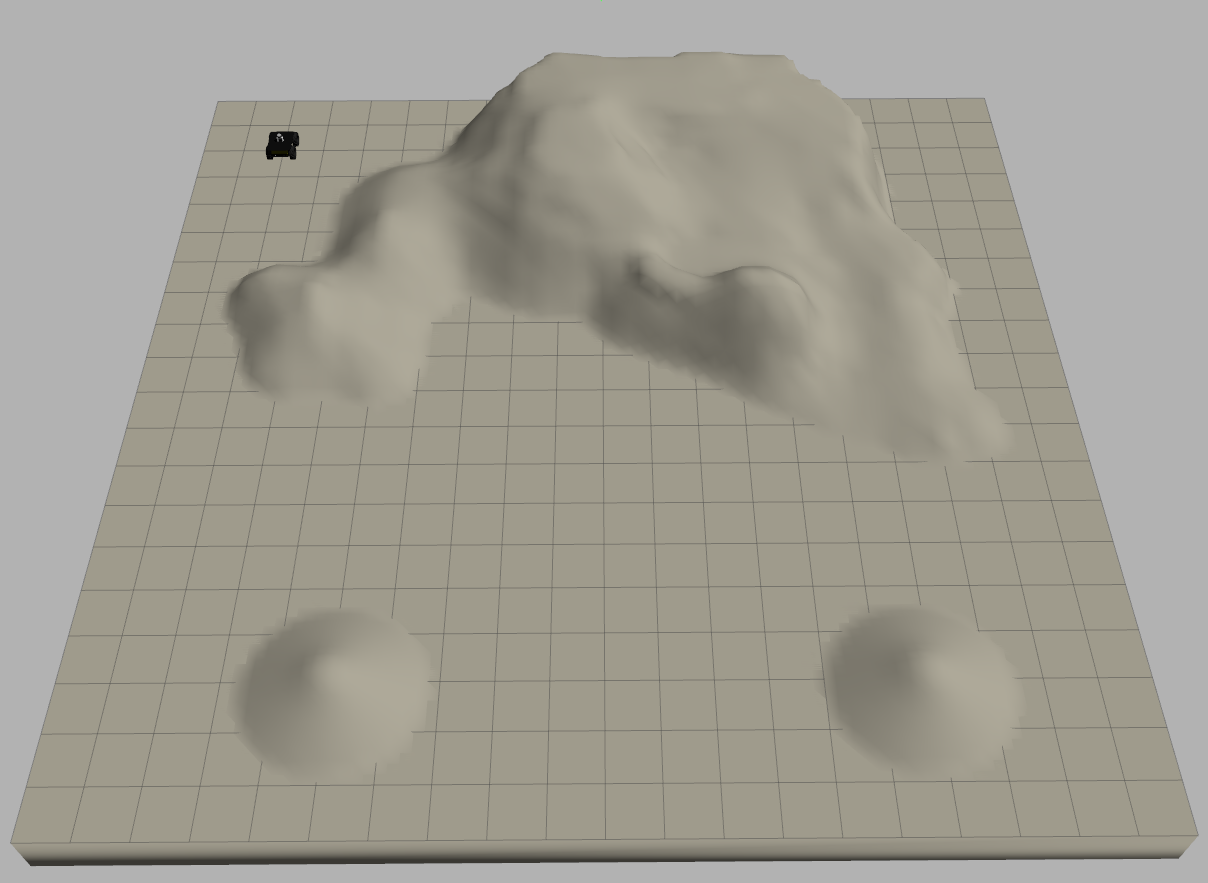} 
  }
\subfloat[\label{fig_sim_b} Topographic Map \textbf{A}]{%
  \includegraphics[width=0.24\textwidth,height=1.4in]{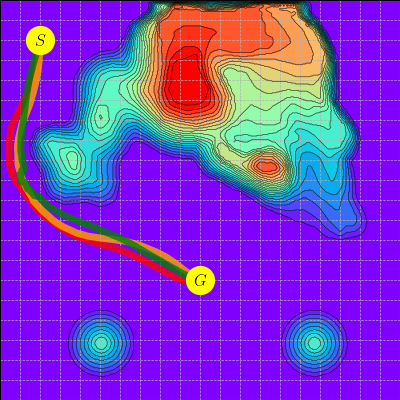} 
  } 
  \vspace{4pt}
\subfloat[\label{fig_sim_c} Gazebo Environment \textbf{B}]{%
  \includegraphics[width=0.24\textwidth,height=1.4in]{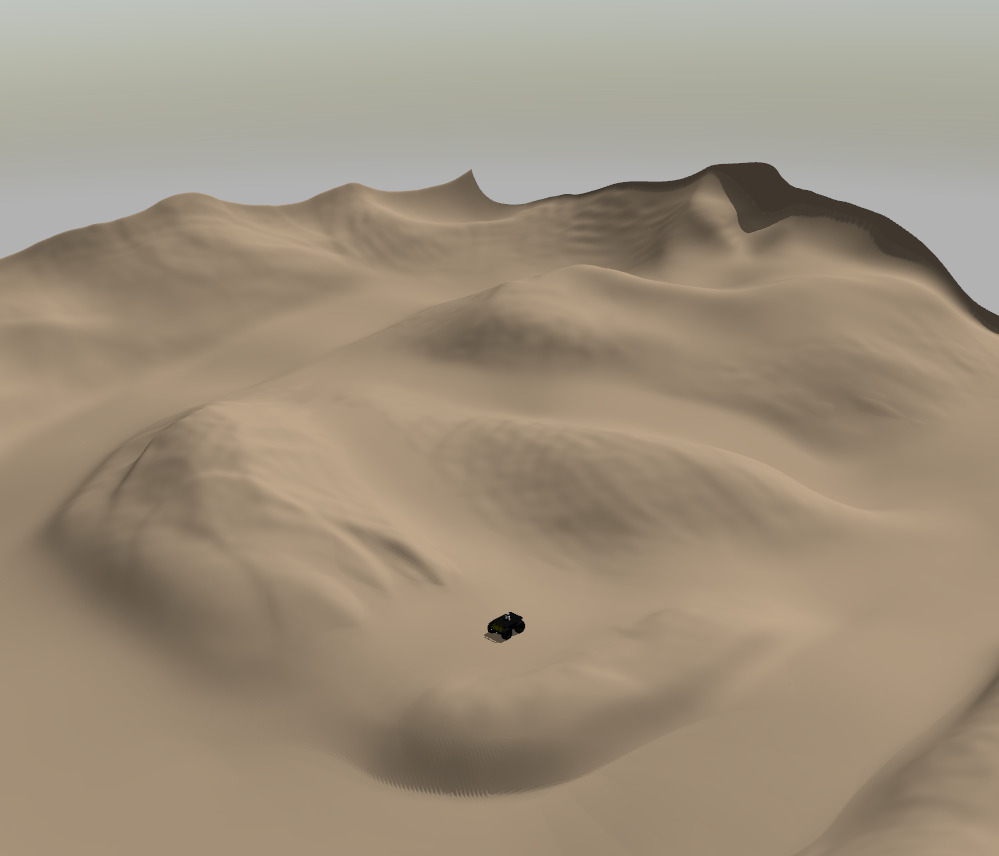} 
} 
\subfloat[\label{fig_sim_d} Topographic Map \textbf{B}]{%
  \includegraphics[width=0.24\textwidth,height=1.4in]{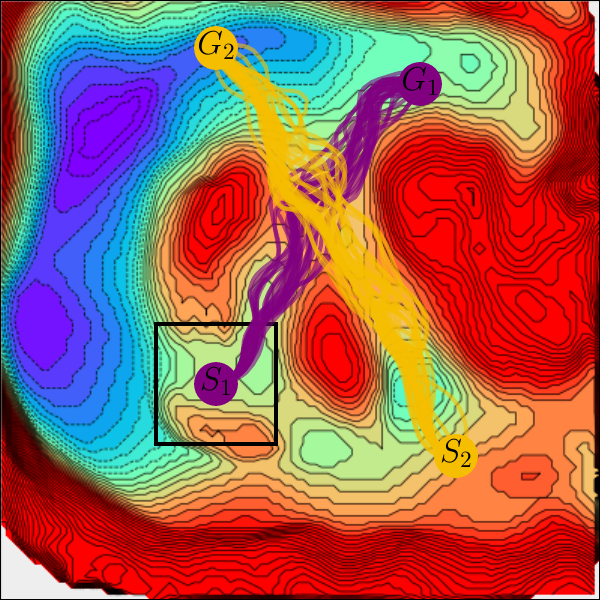} 
} 
  \caption{\small Figure (a) is the Gazebo world for Environment A, provided by~\cite{jian2022putn}, and (b) is the associated topographic map with the safest path achieved by each tested algorithm for task $T_1$. Figure (c) is the Gazebo world for Environment B, and beside it, (d) depicts the topographic map with overlaid paths for all trials run in Tasks 3 and 4. The black box represents the local map $\mathcal{M}$ view, which is visualized in Fig.~\ref{fig_trav}.
  \vspace{-10pt}
  }
  \label{fig_sim_environments}
\end{figure}

During path navigation, each node is sequentially sent to the differential-drive controller~\cite{diffdrive} as a waypoint. As the robot approaches each waypoint, it proceeds to the next one upon reaching a specified proximity to the current waypoint. 
This approach ensures that the robot maintains a consistent speed, as it does not need to slow down or stop during navigation between waypoints.
Similarly, as the robot nears proximity to the local sub-goal, a re-planning phase starts, enabling continuous navigation towards the global goal.



\section{Experimental Design and Results}
\subsection{Simulation Setup} \label{simulation_experiment}
The proposed GP local map built is on top of the GPyTorch \cite{gardner2018gpytorch, Paszke_PyTorch_An_Imperative_2019} library. The RRT* algorithm is implemented in Python. Real-time simulation experiments were conducted in the Gazebo simulator to evaluate the performance of our approach and to compare it to a recent state-of-the-art algorithm for terrain navigation. We selected Plane-fitted Uneven Terrain Navigation (PUTN) Framework~\cite{jian2022putn} as a baseline comparison algorithm. This baseline has already proven its effectiveness and superiority over other methods, and it is open source. During testing, we used two localization modules Advanced-LOAM (ALOAM)~\cite{zhang2014loam}, and precise ground truth provided by a Gazebo plug-in. ALOAM was selected for consistency with the PUTN localization module.


\begin{table*}[t!]
\vspace{5pt}
\centering
\begin{tabular}{|c|c|c|c|c|c|c|c|c|c|}
\hline
 &
   &
   &
   &
   &
   &
  \multicolumn{4}{c|}{Evaluation Criteria} \\ \cline{7-10} 
\multirow{-2}{*}{Env} &
  \multirow{-2}{*}{Task} &
  \multirow{-2}{*}{Start} &
  \multirow{-2}{*}{Goal} &
  \multirow{-2}{*}{Method} &
  \multirow{-2}{*}{Loc.} &
  \multicolumn{1}{c|}{$v_{avg} [\text{m/s}]$} &
  \multicolumn{1}{c|}{$\ell$ [m]} &
  \multicolumn{1}{c|}{$\phi$ [rad]} &
  $\psi$ [rad] \\ \hline
 &
   &
   &
   &
  PUTN &
  AL &
  \multicolumn{1}{c|}{$0.4 \pm 0.01$} &
  \multicolumn{1}{c|}{$20.2 \pm 0.8$} &
  \multicolumn{1}{c|}{$0.146 \pm 0.33$} &
  $\textbf{0.004} \bm{\pm} \textbf{0.002}$ \\
 &
   &
   &
   &
  Ours &
  AL &
  \multicolumn{1}{c|}{$0.45 \pm 0.02$} &
  \multicolumn{1}{c|}{$19.6 \pm 0.6$} &
  \multicolumn{1}{c|}{$\textbf{0.002} \bm{\pm} \textbf{0.002}$} &
  $0.008 \pm 0.17$ \\
\multirow{-3}{*}{A} &
  \multirow{-3}{*}{$T_1$} &
  \multirow{-3}{*}{(-8, 8)} &
  \multirow{-3}{*}{(0, -4)} &
  Ours &
  GT &
  \multicolumn{1}{c|}{$\textbf{0.71} \bm{\pm} \textbf{0.35}$} &
  \multicolumn{1}{c|}{$\textbf{18.0} \bm{\pm} \textbf{0.5}$} &
  \multicolumn{1}{c|}{$0.009 \pm 0.008$} &
  $0.006 \pm 0.004$ \\ \hline
 &
   &
   &
   &
  PUTN [{\color[HTML]{FE0000} Fail}] &
  AL &
  \multicolumn{1}{c|}{$0.3 \pm 0.1$} &
  \multicolumn{1}{c|}{$26.5 \pm 8.8$} &
  \multicolumn{1}{c|}{$0.287 \pm 0.1$} &
  $0.41 \pm 0.11$ \\
 &
  \multirow{-2}{*}{$T_2$} &
  \multirow{-2}{*}{(-7, -7)} &
  \multirow{-2}{*}{(3, 11)} &
  Ours &
  AL &
  \multicolumn{1}{c|}{$\textbf{0.35} \bm{\pm} \textbf{0.04}$} &
  \multicolumn{1}{c|}{$\textbf{38.0} \bm{\pm} \textbf{6.6}$} &
  \multicolumn{1}{c|}{$\textbf{0.37} \bm{\pm} \textbf{0.16}$} &
  {$\textbf{0.52} \bm{\pm} \textbf{0.3}$} \\ \cline{2-10} 
 &
  $T_3$ &
  (-7, -7) &
  (10, 18) &
  Ours &
  GT &
  \multicolumn{1}{c|}{$0.6 \pm 0.02$} &
  \multicolumn{1}{c|}{$33.6 \pm 1.3$} &
  \multicolumn{1}{c|}{$0.32 \pm 0.085$} &
  $0.35 \pm 0.085$ \\ \cline{2-10} 
\multirow{-4}{*}{B} &
  $T_4$ &
  (-13, 13) &
  (21, -7) &
  Ours &
  GT &
  \multicolumn{1}{c|}{$0.51 \pm 0.02$} &
  \multicolumn{1}{c|}{$42.4 \pm 1.4$} &
  \multicolumn{1}{c|}{$0.34 \pm 0.07$} &
  $0.34 \pm 0.05$ \\ \hline
  
\end{tabular}
\caption{\small The table presents simulation evaluation statistics gathered through a series of tasks. The best statistics are in bold. The column labeled Loc. stands for Localization, and the acronyms used are ALOAM (AL) and Ground Truth (GT). More extensive testing was conducted in the more difficult Environment B with 3 Tasks: $T_2$ was a localization task where the baseline failed to reach the goal and our method succeeded. $T_3$ and $T_4$ are robustness testing of our framework, visualized in Fig.~\ref{fig_sim_d}. 
\vspace{-10pt}
    }
    \label{fig_results}
\end{table*}
We assessed the local navigation performance using two environments, A and B. Environment A, sourced from the PUTN GitHub repository \cite{jian2022putn}, features a smooth mountain landscape mainly for testing non-traversable obstacle avoidance seen in Fig.~\ref{fig_sim_a}. In contrast, Environment B presents rugged, steep, and slightly undulating terrain to evaluate the framework under more dynamic conditions shown in Fig.~\ref{fig_sim_c}.

Our evaluation relied on four metrics: average velocity (\(v_{\text{avg}}\)), maximum average robot roll (\(\phi\)), maximum average robot pitch (\(\psi\)), and path length (\(\ell\)), see Table~\ref{fig_results}.
Our goal in the simulation testing is to validate the algorithm's capability to safely navigate uneven terrain abiding by the Clearpath Robotics Husky's \cite{huskyrobot} angular limits of \(0.785 \, \text{rad}\) for pitch and \(0.524 \, \text{rad}\) for roll.
We selected four path objectives, or Tasks across environments A and B, to evaluate various facets of our algorithm's performance. In Task 1, denoted as $T_1$ and simulated in environment A as depicted in Fig.~\ref{fig_sim_a}.
We conducted multiple trials using different localization modules for our proposed method and the baseline. 
In environment B, we conducted three distinct tasks, labeled \(T_2\), \(T_3\), and \(T_4\). The initial and final coordinates for each test are outlined in Table~\ref{fig_results}.
It is important to note that the baseline algorithm could not successfully navigate any of the paths in environment B, failing in all tasks even with parameter tuning. For details on parameter configuration for our method, please refer to our GitHub. The tasks tested were designed to assess our algorithm's efficacy and safety across diverse and challenging navigation scenarios.

\subsection{Simulation Results} \label{sec_sim_results}
\begin{figure} 
  \includegraphics[width=0.24\textwidth,height=1.2in]{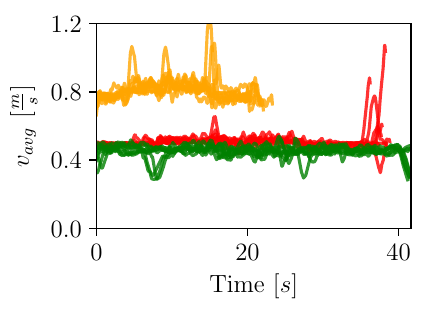} 
  \includegraphics[width=0.24\textwidth,height=1.2in]{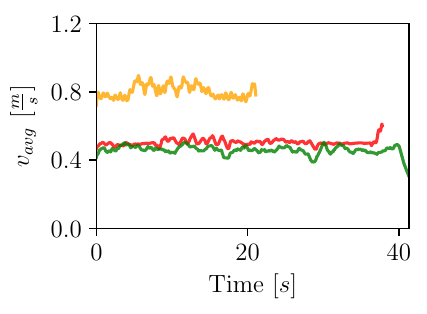} 
  \vspace{0pt}
  \includegraphics[width=0.48\textwidth,height=0.15in]{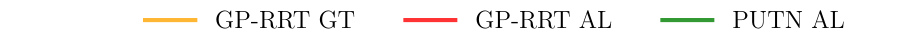} 
  \caption{\small 
    Environment A velocity graphs. The left graph displays the velocities for 10 trials for each method tested in $T_1$. AL and GT stand for ALOAM and Ground Truth respectively. The right graph displays a single velocity for the safest path taken displayed in Fig.~\ref{fig_sim_b}. \vspace{-10pt}
  }
  \label{fig_map1_graph}
\end{figure}

An overview of our simulation statistics is displayed in Table~\ref{fig_results}. In Environment A, when using ALOAM for localization, both our proposed and the baseline (PUTN) algorithms have approximately equivalent behavior when considering speed and safety. In Task 1, the average velocity PUTN performed  was $0.4 \pm 0.01 \text{m/s}$, and ours was $0.45 \pm 0.02  \text{m/s}$, see Fig.~\ref{fig_map1_graph}.
The roll and pitch values are also consistent between both algorithms, with PUTN having a slightly higher average max roll of $0.146 \pm 0.33 \text{rad}$ compared to GP-RRT of $0.002 \pm 0.002 \text{rad}$.
The pitch angles are insignificant as the terrain of this map predominantly consists of flat ground with minor variations in elevation.
However, an advantage of our algorithm was finding the shortest path within this scenario, with PUTN averaging a path length of $20.2 \pm 0.8 \text{m}$ and our algorithm having a path length of $19.6 \pm 0.6 \text{m}$. 
During testing, we observed that, despite PUTN being a map-based algorithm, it engaged in planning routes into local minima situated beside the non-traversable hill in half of the trials. Our map-less method successfully avoided this during local planning, because we categorized leaf nodes as edges into non-traversable terrain, separating them from possible planning objectives. 
When using ground truth localization in this environment, our method improved its performance in speed and finding the optimal path when compared to our method using ALOAM. The average speed increased to $0.71 \pm 0.35 \text{m/s}$, and the path length also reduced to $18.0 \pm 0.5 \text{m}$. 

\begin{figure} 
\centering
\includegraphics[width=0.23\textwidth,height=1.2in]{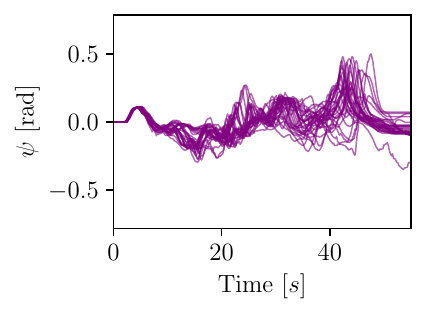} 
\includegraphics[width=0.23\textwidth,height=1.2in]{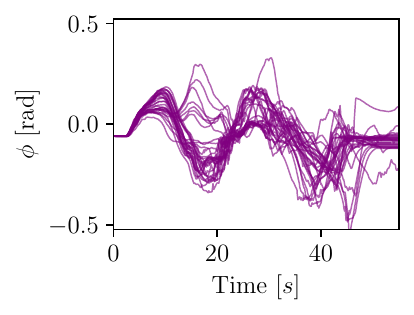}
  \vspace{0pt}
\includegraphics[width=0.23\textwidth,height=1.2in]{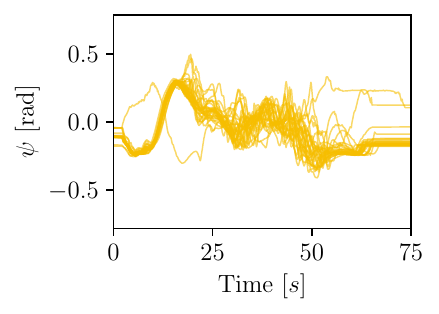} 
\includegraphics[width=0.23\textwidth,height=1.2in]{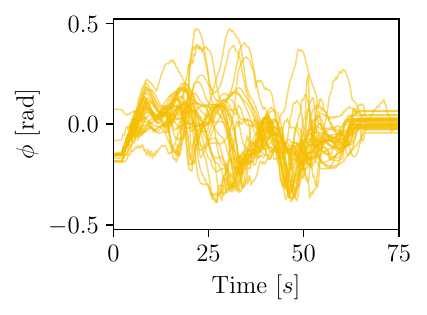} 
\vspace{0pt}
  \includegraphics[width=0.48\textwidth,height=0.15in]{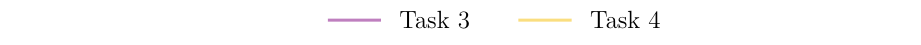}
\caption{\small 
    The graphs in this figure represent the roll and pitch data for \textbf{our proposed method} during the 30 trials for each task in Environment B. The top two graphs showcase the pitch and roll of Task 3, while the ones below depict the data from Task 4. \vspace{-10pt}
  }
  \label{fig_hill1_graph}
\end{figure}

Our algorithm exhibited successful consistent performance across the three navigation tasks in the more challenging Environment B, however, the baseline failed. Since the baseline operates on a map-based method, it necessitates a well-constructed map to facilitate fitting a plane along the generated path. However, the complexity of the environment presented significant challenges, hindering the baseline from establishing a suitable map and ultimately leading to its failure. The results for using ALOAM in that environment are shown in $T_2$ in Table.~\ref{fig_results}. The results demonstrate that our proposed mapless method successfully completed the task, even when utilizing a localization method characterized by a significant amount of noise. The performance metrics gathered from the results are detailed as follows: the average velocity (\(v_{\text{avg}}\)) was recorded as \(0.35 \pm 0.04 \, \text{m/s}\), the total path length (\(\ell\)) was \(38.0 \pm 6.6 \, \text{m}\), the maximum average robot roll (\(\phi\)) was \(0.37 \pm 0.16 \, \text{rad}\), and the maximum average robot pitch (\(\psi\)) was \(0.52 \pm 0.3 \, \text{rad}\).
As seen in Fig.~\ref{fig_hill1_graph}, the roll and pitch values are within the safety specifications for all test cases, confirming our method's robustness to plan trajectories to the objective in an angularly safe manner.
A visualization of all trajectories taken during both tests is displayed in Fig.~\ref{fig_sim_d}.
In the tasks where the ground truth (GT) was used, the average velocity was $0.6 \pm 0.02 \text{m/s}$ on Task 3, and $0.51 \pm 0.02 \text{m/s}$ on Task 4, which shows consistency in terms of speed variation between each of the trials.
The average max roll and pitch for Task 3 were $0.32 \pm 0.085 \text{rad}$ and $0.35 \pm 0.085 \text{rad}$, respectively. In Task 4, we observe similar values $0.34 \pm 0.07 \text{rad}$ and $0.34 \pm 0.05 \text{rad}$. 

The results from the evaluation in both environments offer valuable insights into the local navigation performance of our framework, indicating its strengths in effectively finding and traversing safe paths in an efficient manner. 
For additional information, please refer to the project GitHub for video demonstration and source code.

\section{Conclusion} \label{conclusion}

In this paper, we present a new mapless navigation framework for navigating through rugged terrain. Our proposed approach utilizes a Sparse Gaussian Process (SGP) local map with an optimal Rapidly-Exploring Random Tree (RRT*) planner. 
The GP-based local map serves as the foundation for constructing a traversability map, guiding our navigation planning process.
Our findings, based on extensive simulation tests, reveal the great potential of our framework for enhancing both the safety and efficiency of mapless navigation. 
Our future work will focus on upgrading the controller to optimize its performance and precision.
Additionally, we want to explore the integration of vehicle dynamics into the traversability analysis. Furthermore, we aim to leverage this framework by using learning components such as segmentation.  
These ongoing efforts aim to further advance the capabilities of our navigation framework and contribute to the field of mapless autonomous navigation.


\bibliographystyle{unsrt}
\bibliography{references}

\end{document}